\documentclass{article}

\PassOptionsToPackage{numbers, sort&compress}{natbib}

\usepackage[preprint]{neurips_2026}

\usepackage[utf8]{inputenc} 
\usepackage[T1]{fontenc}    
\usepackage[colorlinks,urlcolor=blue,linkcolor=blue,citecolor=blue]{hyperref}       
\usepackage{url}            
\usepackage{booktabs}       
\usepackage{amsfonts}       
\usepackage{amsmath}        
\usepackage{algorithm}      
\usepackage{algpseudocode}  
\usepackage{nicefrac}       
\usepackage{microtype}      
\usepackage{xcolor}         
\usepackage{url}
\usepackage{enumitem}
\usepackage{graphicx}
\usepackage{bm}
\usepackage{booktabs}
\usepackage{multirow}
\usepackage{colortbl}
\usepackage{wrapfig}
\usepackage{adjustbox}
\usepackage{subcaption}
\usepackage{dsfont}
\usepackage{pifont}
\usepackage{enumitem}
\let\oldding\ding
\renewcommand{\ding}[2][1]{\scalebox{#1}{\oldding{#2}}}
\usepackage{titlesec}
\titlespacing\section{0pt}{1.0ex plus 0.2ex minus 0.1ex}{0.7ex plus 0.1ex}
\titlespacing\subsection{0pt}{0.9ex plus 0.2ex minus 0.1ex}{0.45ex plus 0.1ex}
\titlespacing\subsubsection{0pt}{0.7ex plus 0.2ex minus 0.1ex}{0.35ex plus 0.1ex}
\usepackage{bbding}
\title{ReactSim-Bench: Benchmarking Reactive Behavior World Model Simulation in Autonomous Driving}

\author{%
 Zhiyuan Zhang$^{1,2}$\thanks{This work was partly done when Zhiyuan Zhang interns at  Great Wall Motor.} \And Yanlun Peng$^{2}$ \And Jianing Zhang$^{3}$ \And  Xianda Guo$^{4}$ \AND Zehan Huang$^{1}$ \And Haoran Liu$^{3}$ \And Qifeng Li$^{3}$ \And Shaofeng Zhang$^{5}$ \And Xiaosong Jia$^{3}$\textsuperscript{\Envelope}   \quad\quad\quad Junchi Yan$^{1}$\textsuperscript{\Envelope} 
 \\ \\
 $^{1}$ Sch. of Computer Science \& Sch. of Artificial Intelligence, Shanghai Jiao Tong University\\
$^{2}$ Great Wall Motor \\
$^{3}$ Institute of Trustworthy Embodied AI (TEAI), Fudan University \\
$^{4}$ School of Computer Science, Wuhan University \\
$^{5}$ University of Science and Technology of China \\
\textsuperscript{\Envelope} Correspondence Author
\\
\normalsize{
    \url{https://github.com/Thinklab-SJTU/ReactSim-Bench}
    }
}

\begin{document}

\maketitle

\begin{abstract}
  
Reactive capability is a key property of data-driven behavior world model simulators for autonomous driving simulation systems. With this capability, simulated world agents can respond feasibly to autonomous vehicle (AV) behaviors that differ from the log. However, existing behavior simulation benchmarks do not directly measure reactive capability. They often let the simulator jointly control the AV and surrounding agents and evaluate realism through log similarity or open-loop prediction metrics. In this work, we introduce \textbf{ReactSim-Bench} for evaluating the reactive capability of behavior world model simulation in autonomous driving. We decouple the control of agents and the AV, using AV behaviors that differ from the log and require agents to respond as independent AV inputs. To obtain these AV behaviors, we construct a pipeline that uses an AV planner model to generate candidate behaviors and filters the data using rules and manual verification. Collision metrics, map-based metrics, and kinematic feasibility metrics are used to evaluate the safety and rule compliance of reactive responses. We construct 2,636 test scenarios with three categories and conduct a systematic evaluation of state-of-the-art models across multiple architectures, including Transformer-based, diffusion-based, and next-token-prediction-based models. We further analyze how replan frequency affects performance and provide insights for future studies. 
\end{abstract}

\section{Introduction}

\begin{figure}[t]
  \centering
  \includegraphics[width=\linewidth]{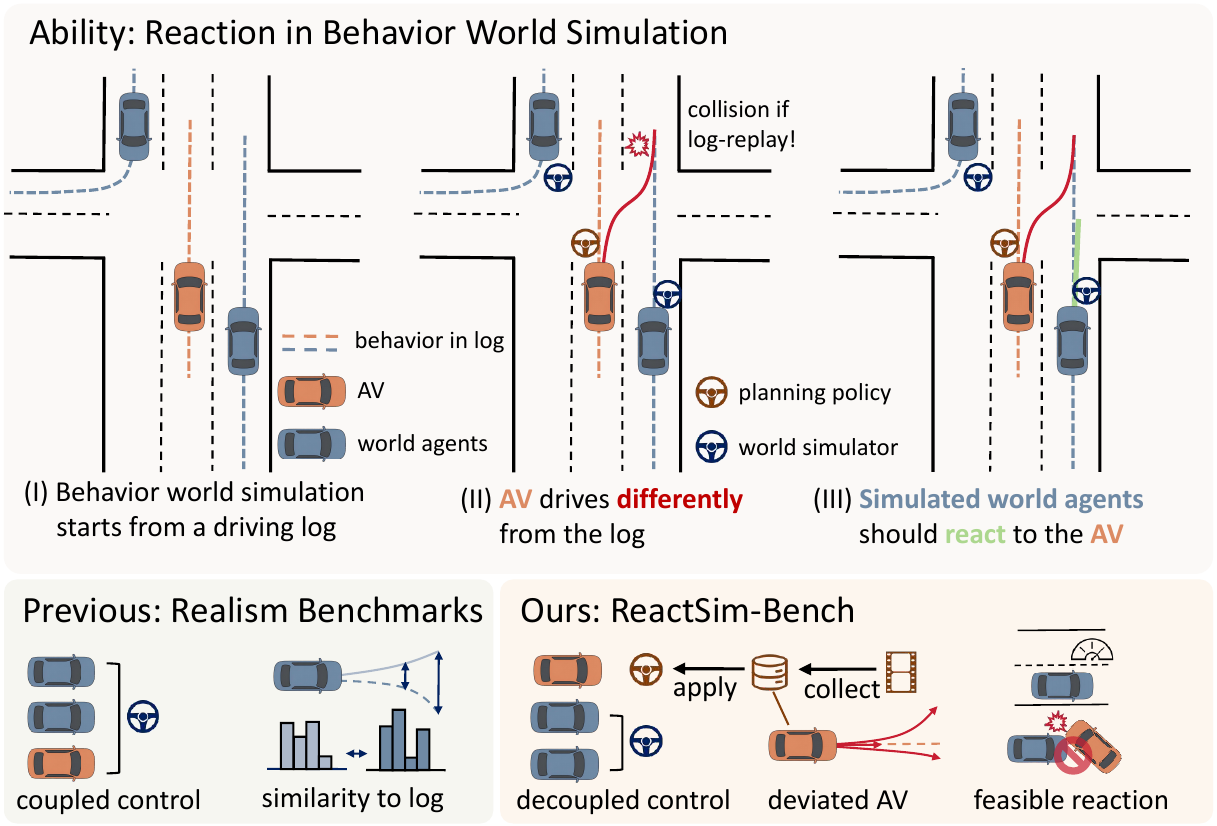}
  \vspace{-12pt}
  \caption{\textbf{Motivation of ReactSim-Bench.} It evaluates whether simulated surrounding agents can react to an AV that deviates from the recorded log, unlike realism benchmarks that couple all vehicles and evaluate log similarity.}
  \label{fig:teaser}
  \vspace{-8pt}
\end{figure}

Simulation is a critical component in the development of autonomous driving (AD) algorithms, for both closed-loop evaluation and reinforcement learning. In recent years, AD simulation systems driven by real-world data have attracted increasing attention~\cite{yang2025drivearena,yang2025resim}. Compared with rule-based simulators, they typically offer higher realism and a smaller sim-to-real gap~\cite{chen2024data}. In such systems, behavior simulation world models control the motion of surrounding vehicles by generating their future behaviors from the environmental context and historical states~\cite{bhattacharyya2019simulating,bhattacharyya2018multi,shin2019randomized,choi2021trajgail,sun2024modelling,sun2024learning,wei2026transformer}.

\textbf{Reactive capability} is a key property of behavior world simulation models, and also one of the main reasons why simulation is needed instead of simple log replay. It answers the question: \textbf{if the autonomous vehicle (AV) controlled by a model to be tested does not strictly follow the recorded driving log, how will surrounding vehicles respond?} As shown in Figure~\ref{fig:teaser} (upper), a behavior simulation model is expected to provide reasonable feedback to the AV, such as following, yielding, and collision avoidance. As simulation rolls forward, the state of the AV increasingly deviates from the logged trajectory, leading to growing distribution shift and posing a major challenge to data-driven behavior world simulation models.  

Although several benchmarks and protocols exist for behavior world simulation, the evaluation of reactive capability remains insufficient. Popular benchmarks such as the Waymo Open Sim Agents Challenge (WOSAC)~\cite{montali2023waymo} mainly evaluate the realism of world simulation. In their setup, both the AV and surrounding agents are controlled by the behavior simulation model, with the objective of making predictions as close as possible to the recorded data. As a result, they do not evaluate interaction with an AV that is not controlled by the behavior simulation model and may deviate from the logged trajectory. Some other evaluation protocols still follow an open-loop paradigm similar to motion prediction, and therefore cannot assess reactive capability~\cite{pronovost2023scenario}.

In this work, we propose \textbf{ReactSim-Bench}, the first benchmark that evaluates the reactive capability of current behavior simulation models, as shown in Figure~\ref{fig:teaser} (lower right). We propose the reactive simulation protocol, which decouples the control of agents and the AV, treating the AV as an independent input to the simulation models. We collect AV behaviors that differ from the log, impose appropriate reactive pressure on agents, and are kinematically feasible as the AV behavior inputs. To obtain such behaviors, we design a data collection pipeline that uses an AV planner model to generate candidate behaviors and filters the data using rules and manual verification. For evaluation, we use collision, map-based, and kinematic metrics to examine the safety and rule compliance of world agents' responses generated by the simulator.

Our benchmark is built on the nuPlan dataset~\cite{caesar2021nuplan}. We construct 2,636 test scenarios and divide them into three categories. We conduct a comprehensive evaluation of state-of-the-art (SOTA) behavior simulation models across multiple paradigms, including classic Transformer-based models (MTR~\cite{shi2022motion}), diffusion-based models (VBD~\cite{huang2026versatile}, CTG~\cite{zhong2023guided}), and next-token-prediction-based models (SMART~\cite{wu2024smart}, CATK~\cite{zhang2025closed}, TrajTok~\cite{zhang2026trajtok}). 
Furthermore, we study how replan frequency affects reactive capability. We hope ReactSim-Bench can provide a practical reference point and encourage future studies on reactive behavior simulation.

Our contributions are four-fold:
\begin{itemize}[leftmargin=10pt, topsep=0pt, itemsep=1pt, partopsep=1pt, parsep=1pt]
    \item We present ReactSim-Bench for evaluating the reactive capability of behavior simulation, which measures how well surrounding-vehicle simulation models handle ego behaviors that deviate from the recorded log in autonomous driving simulators.
    \item We design a pipeline to collect reasonable AV behaviors that differ from the recorded log as AV inputs in ReactSim-Bench.
    \item We systematically evaluate the reactive capability of SOTA models across multiple architectures.
    \item We further analyze the relationship between realism and reactive performance and how replan frequency influences reactive performance.
\end{itemize}

\section{Related work}

\subsection{Behavior Simulation World Models}

Behavior world simulation in autonomous driving generates future behaviors of traffic participants from historical trajectories and scene context. Early simulators mainly rely on log replay or rule-based methods~\cite{behrisch2011sumo,treiber2017intelligent,kesting2007general,kesting2010enhanced}. Data-driven approaches have since been developed for reactive and more realistic simulation~\cite{zheng2020learning,zheng2022traffic,feng2022trafficgen,zhang2023learning,zhang2023trafficbots,zhang2024trafficbots,lin2025revisit,li2023scenarionet,tan2021scenegen,wang2022learning,rowe2025scenario,pourkeshavatz2026autoworld,chitta2024sledge,luo2022gamma}. Recently, diffusion-based methods and next-token-prediction (NTP) based methods are two representative directions for behavior simulation world models.

Diffusion is used in behavior simulation world models to model the distribution of multimodal behaviors and generate controllable behaviors~\cite{chang2024safe,liu2025rolling,yao2025ep,lu2024data,chauhan2025vfsi,jiang2024scenediffuser,pronovost2023scenario,jiang2023motiondiffuser,tan2025scenediffuser++,westny2024diffusion,yang2025drivearena,zhang2024lcsim,liu2025adv,yang2025x,guo2025genesis,cao2025pseudo}. For example, CTG~\cite{zhong2023guided} introduces conditional diffusion with signal temporal logic guidance for controllable traffic generation, and VBD~\cite{huang2026versatile} extends this idea to generalized traffic-agent simulation with scene-consistent interactions. The NTP paradigm is adopted for autoregressive generation~\cite{wang2023multiverse,qian20232nd,zhou2024behaviorgpt,pei2025advancing,peng2025infgen,peng2024improving,chang2025spacer,cornelisse2025building,peng2026scenestreamer,tan2024promptable,chen2025rift,gao2025laser,wang2025llm,you2024bench2drive,yang2025resim,ahmadi2025rlftsim,wang2026learning,song2026unimm}. Trajeglish~\cite{philion2023trajeglish} first models traffic behavior generation as discrete NTP with a data-driven K-disks tokenizer. SMART~\cite{wu2024smart} introduces map and trajectory tokenization and studies architectural scalability, while KiGRAS~\cite{zhao2024kigras} factorizes driving scenes in action space with kinematic transformations. CATK~\cite{zhang2025closed} then addresses the closed-loop OOD problem with supervised fine-tuning, and TrajTok~\cite{zhang2026trajtok} further improves the tokenizer by analyzing coverage, utilization, symmetry, and robustness.

\subsection{Behavior Simulation Protocols and Benchmarks}

Existing behavior-simulation protocols are often defined with individual models and datasets. TrafficSim~\cite{suo2021trafficsim} evaluates generated traffic on the ATG4D dataset~\cite{luo2018fast} using metrics such as scenario collision rate and average displacement error, while BITS~\cite{xu2022bits} reports failure rate and diversity on Lyft~\cite{li2023large} and nuScenes~\cite{caesar2020nuscenes}. These metrics cover realism, safety, and diversity, but their computation details, validation splits, and rollout settings vary, making unified comparison difficult. The Waymo Open Sim Agents Challenge (WOSAC)~\cite{montali2023waymo} is one of the most complete and popular benchmarks for this area. It focuses on realism by computing the negative log-likelihood of multiple features extracted from simulated rollouts against logged driving statistics. It also standardizes protocol details such as the number of rollouts and the replanning frequency.

However, there still lacks benchmark for reactive capability. Most existing benchmarks and protocols predict the AV and surrounding vehicles jointly, which measures whether the simulated scene remains close to the log but does not test the case where the AV trajectory differs from the logged trajectory. Some studies show a few qualitative cases in which some vehicles follow the user-specified behaviors, but they lack a systematic quantitative protocol for reactive ability~\cite{huang2026versatile}.

\section{ReactSim-Bench}

\subsection{Reactive Behavior World Simulation Task}
\label{sec:3.1}

\noindent\textbf{Formulation of behavior world simulation.}
Autonomous driving simulation starts from an initial scenario that contains an HD map $\mathcal{M}$, the past $T_h$ steps, including the current time step, of the AV states $\{ \mathcal{S}^{\text{AV}}_{-T_h}, \ldots, \mathcal{S}^{\text{AV}}_{0} \}$, and the world agent states $\{ \mathcal{S}^{\text{agents}}_{-T_h}, \ldots, \mathcal{S}^{\text{agents}}_{0} \}$. At each step $t>0$, the AV policy $p$ controls the next AV state based on the logged history $\mathcal{S}_{-T_h:0}=[\mathcal{S}^{\text{AV}}_{-T_h:0},\mathcal{S}^{\text{agents}}_{-T_h:0}]$ and the already executed states $\mathcal{S}'_{0:t}=[\mathcal{S}'^{\text{AV}}_{0:t},\mathcal{S}'^{\text{agents}}_{0:t}]$:
\begin{equation}
\mathcal{S}'^{\text{AV}}_{t+1}=p(\mathcal{M},[\mathcal{S}_{-T_h:0},\mathcal{S}'_{0:t}]).
\end{equation}

The simulator $q$ updates the states of the world agents in the environment using the same historical information:
\begin{equation}
\mathcal{S}'^{\text{agents}}_{t+1}=q(\mathcal{M},[\mathcal{S}_{-T_h:0},\mathcal{S}'_{0:t}]).
\end{equation}

After the AV policy and the simulator complete their respective updates, the simulation advances to time step $t+1$. Repeating this process yields an autoregressive closed-loop rollout.

\noindent\textbf{Importance of reaction in behavior world simulation.}
Behavior world simulation is usually grounded in a real driving log, which contains both the pre-simulation states $\mathcal{S}_{-T_h:0}$ and the future expert states $\mathcal{S}_{0:T_f}=[\mathcal{S}^{\text{AV}}_{0:T_f},\mathcal{S}^{\text{agents}}_{0:T_f}]$ over the next $T_f$ steps. We usually expect the AV policy to behave similarly to the expert, i.e., the AV rollout $\mathcal{S}'^{\text{AV}}_{0:T_f}$ remains close to the logged AV trajectory $\mathcal{S}^{\text{AV}}_{0:T_f}$. If the AV exactly reproduces the expert states, a simulator can simply perform log replay and output the logged surrounding agent states $\mathcal{S}^{\text{agents}}_{0:T_f}$, which are closest to real driving behavior. However, AV policies often deviate from the expert. Therefore, a behavior simulator must produce reasonable responses from surrounding agents when the AV trajectory differs from the recorded log.

\noindent\textbf{Reactive task in ReactSim-Bench.} 
The reactive simulation task in ReactSim-Bench is shown in Algorithm~\ref{alg:reactsim}. We provide AV behaviors that differ from the logged ones $\mathcal{S}^{\text{deviated AV}}_{0:T_f}$ as the AV policy in the behavior world simulation process described above. Notably, the neural network in simulator $q$ is not required to infer the next state at every step. It can infer several states at a time and apply them step by step.

\begin{figure}[t]
  \centering
  \vspace{-12pt}
  \includegraphics[width=0.70\linewidth]{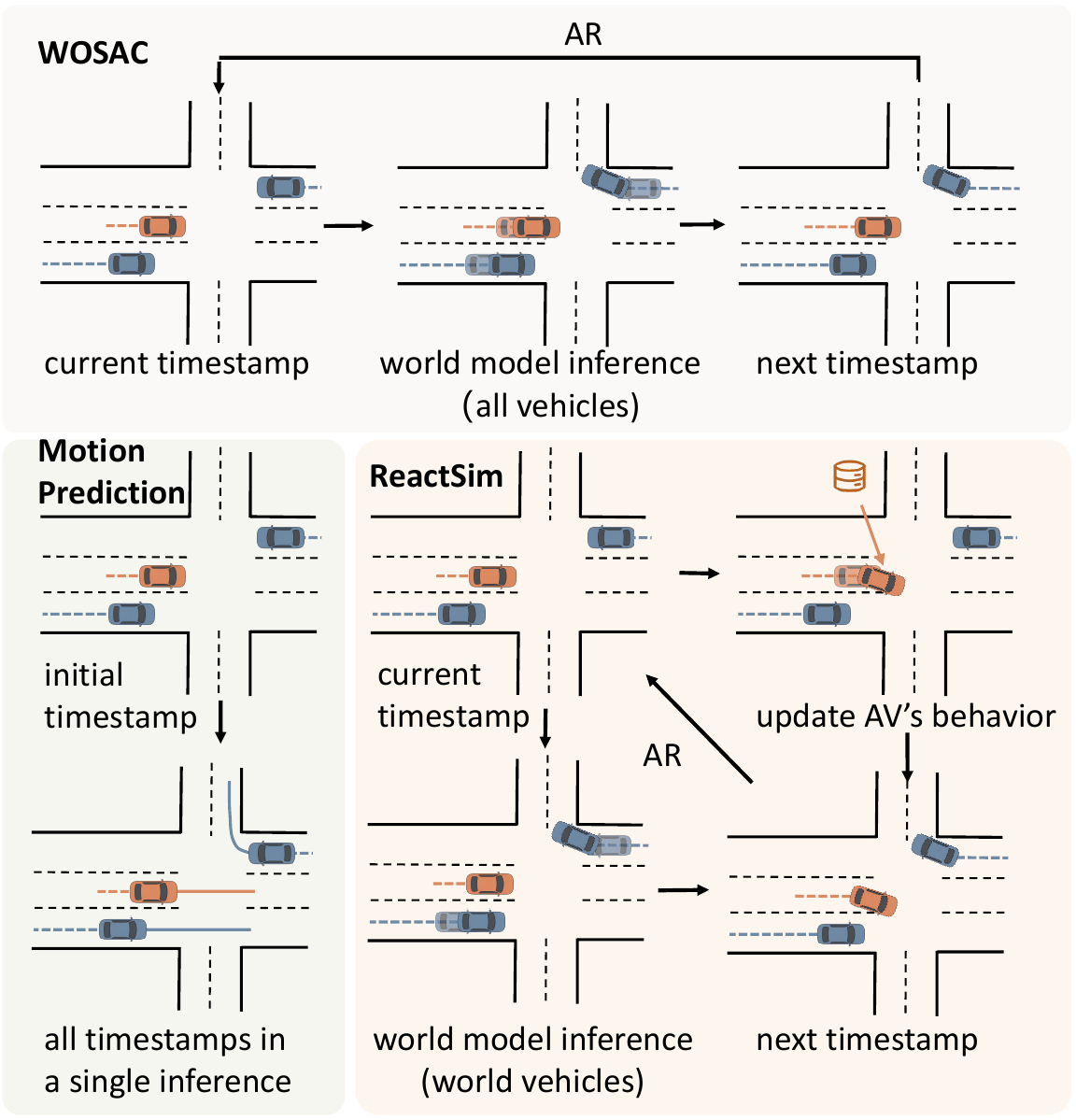}
  \caption{\textbf{Protocol comparison.} WOSAC controls all vehicles autoregressively for realism evaluation, motion prediction predicts all timestamps in one pass, and ReactSim-Bench decouples AV and agent behavior in an autoregressive rollout.}
  \label{fig:protocol}
  \vspace{-4pt}
\end{figure}

\begin{algorithm}[t]
\caption{ReactSim-Bench: Closed-Loop, Agent-AV Decoupled Inference, Reactive Task}
\label{alg:reactsim}
\begin{algorithmic}[1]
\Statex \textbf{Input:} HD map $\mathcal{M}$; logged history $\mathcal{S}_{-T_h:0}$; fixed deviated AV trajectory $\mathcal{S}^{\text{deviated AV}}_{0:T_f}$; behavior simulator $q$; horizon $T_f$.
\State Initialize $\mathcal{S}'^{\text{AV}}_{0}\gets \mathcal{S}^{\text{deviated AV}}_{0}$ and $\mathcal{S}'^{\text{agents}}_{0}\gets \mathcal{S}^{\text{agents}}_{0}$.
\For{$t=0,\ldots,T_f-1$}
    \State $\mathcal{S}'^{\text{AV}}_{t+1}\gets \mathcal{S}^{\text{deviated AV}}_{t+1}$ \Comment{The AV is an external input, not controlled by $q$.}
    \State $\mathcal{S}'^{\text{agents}}_{t+1}\gets q(\mathcal{M},[\mathcal{S}_{-T_h:0},\mathcal{S}'^{\text{AV}}_{0:t+1},\mathcal{S}'^{\text{agents}}_{0:t}])$ \Comment{Agents react to the realized AV state.}
\EndFor
\State \Return $[\mathcal{S}'^{\text{AV}}_{0:T_f},\mathcal{S}'^{\text{agents}}_{0:T_f}]$.
\end{algorithmic}

\end{algorithm}

\begin{algorithm}[t]
\caption{WOSAC: Closed-Loop, Agent-AV Coupled Inference, Realism Task}
\label{alg:wosac}
\begin{algorithmic}[1]
\Statex \textbf{Input:} HD map $\mathcal{M}$; logged history $\mathcal{S}_{-T_h:0}$; logged future $\mathcal{S}_{1:T_f}$; behavior simulator $q$; horizon $T_f$.
\State Initialize $\mathcal{S}'_{0}\gets [\mathcal{S}^{\text{AV}}_{0},\mathcal{S}^{\text{agents}}_{0}]$.
\For{$t=0,\ldots,T_f-1$}
    \State $\mathcal{S}'_{t+1}\gets q(\mathcal{M},[\mathcal{S}_{-T_h:0},\mathcal{S}'_{0:t}])$ \Comment{$q$ controls both the AV and surrounding agents.}
\EndFor
\State \Return $\mathcal{S}'_{1:T_f}$.
\end{algorithmic}
\end{algorithm}

\noindent\textbf{Differences between ReactSim-Bench and previous realism benchmarks.}
Existing behavior world simulation benchmarks, such as WOSAC, also formulate autonomous driving simulation as stepwise inference and execution by an AV policy and a simulator. However, their evaluation protocols hand both the AV and surrounding agents to the simulator:
\begin{equation}
\mathcal{S}'_{t+1}=q(\mathcal{M},[\mathcal{S}_{-T_h:0},\mathcal{S}'_{0:t}]).
\end{equation}

As summarized in Algorithm~\ref{alg:wosac} and Figure~\ref{fig:protocol}, \textbf{this process in WOSAC is an autoregressive version of motion prediction}. The realism metrics evaluate the similarity between the rollouts and the log, and their goal is also similar to motion prediction. Under this protocol, a model can obtain strong scores by fitting the log as closely as possible, making $\mathcal{S}'_{t+1}=[\mathcal{S}'^{\text{AV}}_{t+1},\mathcal{S}'^{\text{agents}}_{t+1}]$ close to $[\mathcal{S}^{\text{AV}}_{t+1},\mathcal{S}^{\text{agents}}_{t+1}]$, while the reactive capability required in practical simulation is ignored. Moreover, when the simulator controls both the AV and surrounding agents, its predicted AV state is guaranteed to be realized in the future rollout, allowing it to coordinate the AV and other agents to avoid potential collisions. In practical simulation, the AV action can differ from the simulator's prediction, making collisions and other abnormal interactions more likely.

\subsection{Deviated AV Data for ReactSim-Bench}

\subsubsection{Data Requirements}

\begin{figure}[t]
  \centering
  \vspace{-10pt}
  \includegraphics[width=1\linewidth]{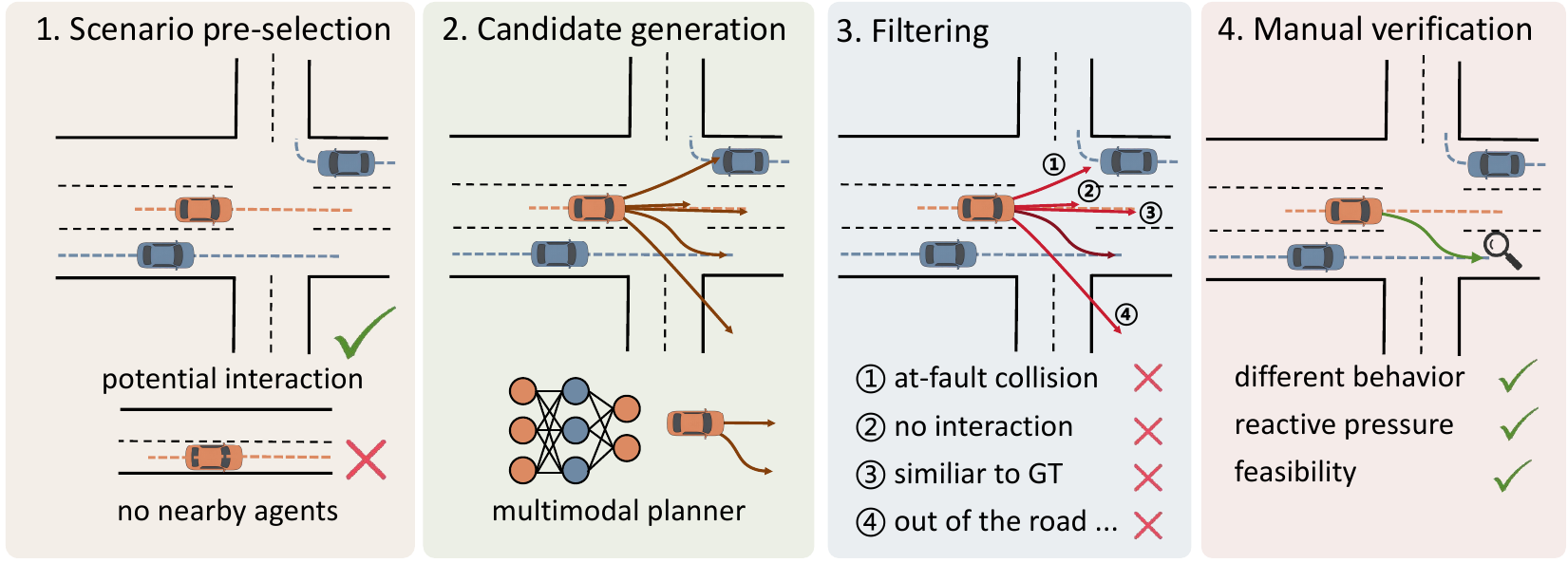}
  \vspace{-18pt}
  \caption{\textbf{Data collection pipeline.} We pre-select interactive scenarios, generate candidate AV trajectories with a multimodal planner, filter invalid candidates, and manually verify the final deviated AV behaviors.}
  \label{fig:data_collection}
\end{figure}

The AV trajectories used to test reactive capability are expected to satisfy three criteria:

\begin{itemize}[leftmargin=10pt, topsep=0pt, itemsep=1pt, partopsep=1pt, parsep=1pt]
  \item \textbf{Behavioral deviation from the log.} The AV behavior should differ sufficiently from the recorded trajectory to evaluate agent reactivity beyond log replay.
  \item \textbf{Reactive pressure.} The deviated AV trajectory should affect the logged motion of surrounding agents, for example by inducing a collision or a small time-to-collision (TTC) under log replay. This makes log replay infeasible and requires surrounding agents to react to avoid potential collisions. Meanwhile, the induced conflict should be avoidable through reasonable agent behavior.
  \item \textbf{Feasibility.} The deviated AV behavior should be kinematically feasible and satisfy map-related constraints, such as staying within the drivable area. 
\end{itemize}

\subsubsection{Data Collection Pipeline}

The most direct way to get such data is to manually select scenarios and draw all future AV states that meet the above requirements, but it requires a lot of human effort. Thus, we design a data collection pipeline with model-based candidate generation and rule-based filtering, as shown in Figure~\ref{fig:data_collection}. The pipeline has four steps:

\begin{itemize}[leftmargin=10pt, topsep=0pt, itemsep=1pt, partopsep=1pt, parsep=1pt]
  \item \textbf{Step 1: Scenario pre-selection.} We first prescreen scenarios in the dataset. Based on the number, distance, and TTC of surrounding vehicles around the AV, we select scenarios with potential AV-agent interactions.
  \item \textbf{Step 2: Candidate generation.} We use a pretrained planner~\cite{zheng2025diffusion} with strong multimodal capability as the AV policy and run inference in the pre-selected scenarios. Specifically, we use Diffusion Planner for this step. The resulting trajectories are treated as candidates. This converts the labor-intensive trajectory drawing process into a trajectory selection process.
  \item \textbf{Step 3: Filtering.} We filter invalid trajectories with a set of rules to reduce the workload of subsequent manual selection. Specifically, we remove trajectories that are too close to the log according to average displacement error (ADE), leave the drivable area, violate kinematic constraints, lack sufficient interaction with other vehicles according to TTC, or actively collide with stationary or slow-moving vehicles.
  \item \textbf{Step 4: Manual verification.} We further verify the filtered trajectories manually to ensure that they satisfy the requirements above. If multiple candidate trajectories in a scenario meet the requirements, we select one of them. If all the candidates are invalid, the scenario is discarded.
\end{itemize}

\subsubsection{Deviation Taxonomy}

\begin{figure}[t]
  \centering
  \includegraphics[width=0.8\linewidth]{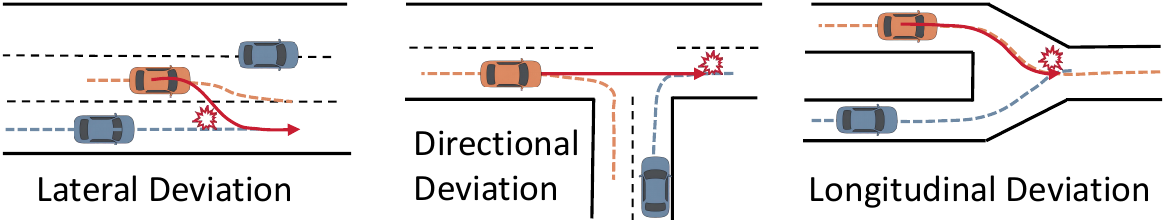}
  \caption{\textbf{Typical cases in each deviation taxonomy.} We group collected AV behaviors into lateral, directional, and longitudinal deviations.}
  \label{fig:classify}
\end{figure}

We categorize collected AV behaviors according to the dominant source of deviation from the log, which reflects the different forms of reactive pressure shown in Figure~\ref{fig:classify}.   

\begin{itemize}[leftmargin=10pt, topsep=0pt, itemsep=1pt, partopsep=1pt, parsep=1pt]
  \item \textbf{Directional deviation.} In this category, the AV drives in a different direction from the log, such as taking a different turning direction or entering a different lane connector. 
  \item \textbf{Lateral deviation.}  In this category, the AV keeps a similar overall driving direction to the log but drives a shifted lateral position,  such as a lane change, a different merge position, or a slightly displaced path through an interaction region. 
  \item \textbf{Longitudinal deviation.} In this category, the AV follows a spatial path similar to the logged trajectory, but its temporal profile differs, such as moving faster, slowing down, or stopping. 
\end{itemize}

When a trajectory exhibits multiple deviation cues, we assign it to the category that best explains the interaction conflict induced by the deviated AV trajectory.

\subsubsection{Data Analysis}

\begin{figure}[b]
  \centering
  \includegraphics[width=1.0\linewidth]{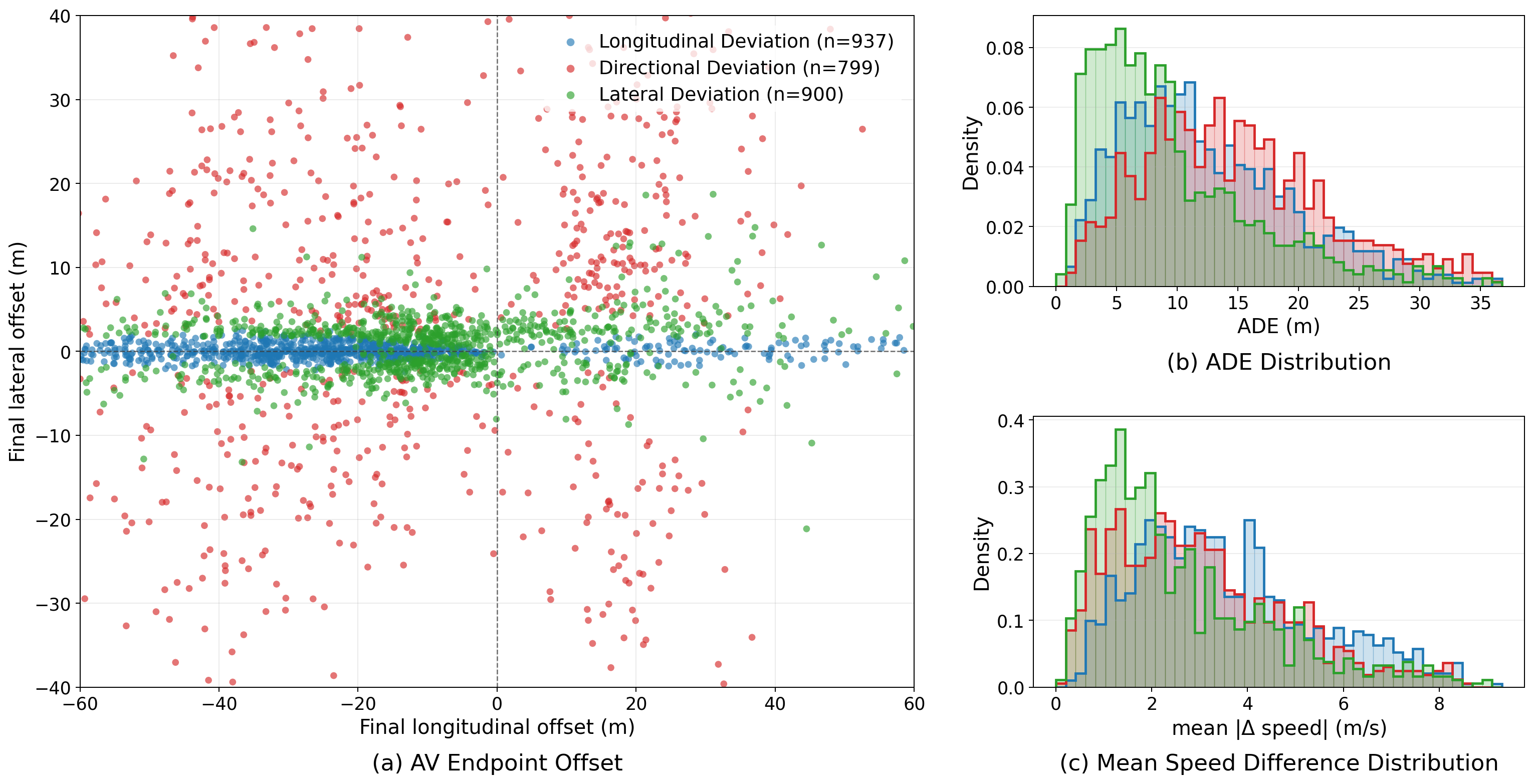}
  \caption{\textbf{Distribution of collected AV deviations.} (a) Endpoint offsets by deviation type. (b) ADE and (c) mean speed difference between deviated and logged AV trajectories.}
  \label{fig:distribution}
\end{figure}

\noindent\textbf{Data distribution.}
We construct the test set of ReactSim-Bench on the nuPlan~\cite{caesar2021nuplan} dataset. The test set contains 2,636 scenarios with deviated AV behaviors, including 937 longitudinal deviations, 799 directional deviations, and 900 lateral deviations. Figure~\ref{fig:distribution}(a) visualizes the endpoint offset of each collected AV trajectory relative to the logged AV trajectory. The endpoint distribution is consistent with the taxonomy above: longitudinal deviations concentrate near zero lateral offset, lateral deviations show clear lateral displacement, and directional deviations spread over a wider spatial range. 
Figures~\ref{fig:distribution}(b) and~\ref{fig:distribution}(c) further show that the collected AV behaviors differ substantially from the recorded log. The ADE distributions are separated from zero and extend over a wide range, while the mean speed-difference distributions show non-negligible temporal deviations from the logged AV motion. Different categories emphasize different deviation sources, with directional cases tending to have larger spatial differences and longitudinal cases more directly affecting speed profiles. 

\noindent\textbf{Reactive pressure.} To analyze the reactive pressure, we replay the logged surrounding agent behaviors while replacing the AV future behavior with the collected data, and calculate the AV-agent collision count and TTC. Log replay produces at least one AV-agent collision in 83.46\% of scenarios and risky interactions with a minimum TTC below $\tau_{\mathrm{TTC}}=\text{0.5}$~s in 98.14\% of scenarios. This means the world agents in our test scenarios should react to the AV to avoid unsafe interactions, rather than just replay the log.

\subsection{Metrics}

We aim to evaluate whether surrounding agents can respond reasonably to a deviated AV input in the reactive setting. To this end, we select several evaluation dimensions that are commonly used in autonomous-driving evaluation~\cite{gulino2023waymax,caesar2021nuplan} for measuring driving safety and feasibility, and adapt their computation to our protocol. The resulting metrics cover AV-agent safety, agent-agent safety, map and driving-direction compliance, and kinematic feasibility:

\begin{itemize}[leftmargin=10pt, topsep=0pt, itemsep=1pt, partopsep=1pt, parsep=1pt]
\item \textbf{Agent-AV collision count} reports the average number of surrounding agents that collide with the AV, reflecting whether agents can respond to AV behavior to avoid potential collisions. For each scenario, we first identify all agents that collide with the AV at any time step. For each such agent, at its first collision time with the AV, we determine whether the agent is at fault by their position and heading. The agent is counted only if it is at fault.

\item \textbf{Agent-AV risky TTC count} is a stricter counterpart to the average Agent-AV collision count. We compute the TTC between each surrounding agent and the AV. If the TTC falls below a threshold at any time step, the agent is marked as risky. The final metric is the average risky-agent count over all scenarios.

\item \textbf{Agent-agent collision rate} computes the fraction of world agents involved in collisions with other world agents at each step.

\item \textbf{Off-road rate} computes the fraction of timestamps at which a vehicle leaves the drivable area. We only evaluate vehicles that are initially in the drivable area.

\item \textbf{Driving-direction violation rate} computes the fraction of vehicles that violate the driving direction of their current road segment, such as driving against traffic.

\item \textbf{Kinematic infeasibility rate} estimates each vehicle's acceleration and steering curvature from consecutive states using an inverse kinematic model and checks whether they remain within predefined thresholds. We report acceleration and steering-curvature infeasibility rates separately.
\end{itemize}

Detailed metric computation rules are provided in Appendix~\ref{app:metric_details}.

\section{Experiments}

\subsection{Datasets \& Baselines}

Our benchmark is built on the nuPlan~\cite{caesar2021nuplan} dataset, which contains more than 1,000 hours of driving data. We repartition the data into a training set with 642,640 scenarios and a test set with 2,636 scenarios. Each scenario lasts 9 seconds, consisting of 1 second of history and 8 seconds of future rollout. The sampling frequency is 10~Hz, resulting in 91 frames per scenario. We select the 64 agents closest to the AV at the current time step, i.e., the 10th frame, as prediction targets. For map inputs, we use four types of map elements, with the input range following the original setting of each baseline. Our baselines include:

\begin{itemize}[leftmargin=10pt, topsep=0pt, itemsep=1pt, partopsep=1pt, parsep=1pt]
\item \textbf{MTR}~\cite{shi2022motion} is a classic Transformer-based motion prediction model. It uses global intention localization and local movement refinement, and achieves strong performance in open-loop trajectory prediction for surrounding agents. We use it as a baseline that directly applies a motion prediction model to behavior simulation.

\item \textbf{CTG}~\cite{zhong2023guided} is one of the earliest works that applies diffusion models to behavior simulation. It takes rasterized environmental and historical information as input and uses signal temporal logic (STL) as guidance to generate realistic and controllable behaviors.

\item \textbf{VBD}~\cite{huang2026versatile} uses a behavior prior predictor to predict a goal point for each agent, and then uses the predicted goals as guidance for a diffusion decoder to generate diverse surrounding-agent behaviors.

\item \textbf{SMART}~\cite{wu2024smart} follows the next-token prediction paradigm. It discretizes agent trajectories into a vocabulary and generates trajectories autoregressively, winning WOSAC 2024.

\item \textbf{CATK}~\cite{zhang2025closed} builds on the NTP paradigm and proposes Closed-Loop Supervised Fine-Tuning, which further improves performance through post-training.

\item \textbf{TrajTok}~\cite{zhang2026trajtok} studies vocabulary design under the NTP paradigm and proposes a trajectory vocabulary with strong coverage, utilization, symmetry, and robustness, winning WOSAC 2025.
\end{itemize}

The above methods were originally implemented on Waymo or nuScenes. We implement them on the nuPlan dataset for evaluation under our benchmark.

\subsection{Main Results}

\begin{table}[b]
  \caption{\textbf{Main results on ReactSim-Bench.} All methods are evaluated with a replan rate of 2~Hz. A-AV Coll. Count denotes Agent-AV collision count, A-AV risky Count denotes Agent-AV risky TTC count, A-A Coll. denotes Agent-Agent collision rate, Dir. denotes driving-direction violation rate, Acc. denotes acceleration infeasibility rate, and Steer. denotes steering-curvature infeasibility rate.}
  \label{tab:main_results}
  \centering
  \small
  \setlength{\tabcolsep}{3.5pt}
  \resizebox{0.99\linewidth}{!}{
  \begin{tabular}{lccccccc}
    \toprule
    Method & \textbf{A-AV Coll. Count} $\downarrow$ & \textbf{A-AV risky Count} $\downarrow$ & A-A Coll. $\downarrow$ (\%) & Offroad $\downarrow$ (\%) & Dir. (\%) $\downarrow$& Acc. (\%) $\downarrow$ & Steer. (\%) $\downarrow$ \\
    \midrule
    Log Replay                 & 0.9829 & 1.5380 & 2.25 & 0.18 & 0.80 & 0.16  & 2.51 \\
    \midrule
    \rowcolor{black!5} MTR     & 0.1457 & 0.5819 & 3.29 & 2.67 & 2.83 & 0.64  & 14.29 \\
    CTG                        & 0.6195 & 0.9476 & 4.88 & 2.95 & 2.10 & 10.87 & 7.08  \\
    \rowcolor{black!5} VBD     & 0.2276 & 0.4711 & 3.19 & 1.03 & 2.35 & 0.01  & 0.18  \\
    SMART                      & 0.1419 & 0.3976 & 2.23 & 0.68 & 1.09 & 9.74  & 4.83  \\
    \rowcolor{black!5} CATK    & 0.1426 & 0.4029 & 2.22 & 0.69 & 1.13 & 10.25 & 5.02  \\
    TrajTok                    & 0.1407 & 0.4173 & 2.23 & 0.61 & 1.03 & 3.23  & 3.93  \\
    \bottomrule
  \end{tabular}}
\end{table}

Table~\ref{tab:main_results} reports all metrics on ReactSim-Bench, and Table~\ref{tab:realism_reactive} compares these methods under our Reactive Protocol and the prior Realism Protocol. We make the following findings:

\textbf{Behavior world simulation models have a certain level of reactive capability.} When the AV behavior differs from the log, log replay produces many collisions and unsafe interactions, while all learned models substantially reduce the A-AV collision count and risky TTC count. This indicates that our test scenarios impose reactive pressure while still remaining solvable.

\textbf{Better realism does not necessarily lead to better reactive performance.} For example, MTR has worse ADE and kinematic likelihood than CTG under the realism protocol, but achieves lower A-AV collision and risky TTC counts under the reactive protocol. Similarly, CATK improves the realism metrics over SMART but has slightly worse reactive metrics. Realism is an important criterion for evaluating behavior world simulation models, but it cannot represent every aspect of model performance.

\textbf{Learning strong reactive capability from fitting the log is difficult.} SMART, CATK, and TrajTok fit the logged trajectories well under the realism protocol, as reflected by their low ADE and very low A-AV collision counts. However, their A-AV collision counts increase markedly under the Reactive Protocol. Imitation-learning paradigms rely on fitting the log and therefore face an out-of-distribution problem when responding to AV trajectories that differ from the log. Although diffusion and next-token-prediction paradigms provide stronger probabilistic modeling than direct regression, understanding and responding to the behavior of other vehicles remains challenging.

\begin{table}[t]
  \caption{\textbf{Results in realism and reactive evaluation protocols.} Kin. Lik. denotes kinematic likelihood in WOSAC.}
  \label{tab:realism_reactive}
  \centering
  \small
  \setlength{\tabcolsep}{4.0pt}
  \resizebox{0.99\linewidth}{!}{
  \begin{tabular}{lcccccc}
    \toprule
    & \multicolumn{2}{c}{Reactive Protocol} & \multicolumn{4}{c}{Realism Protocol} \\
    \cmidrule(lr){2-3}\cmidrule(lr){4-7}
    Method & A-AV Coll. Count $\downarrow$ & A-AV risky Count $\downarrow$ & A-AV Coll. Count $\downarrow$ & A-AV risky Count $\downarrow$ & ADE (m) $\downarrow$ & Kin. Lik. $\uparrow$ \\
    \midrule
    \rowcolor{black!5} MTR     & 0.1457 & 0.5819 & 0.0781 & 0.5353 & 2.5498 & 0.2379 \\
    CTG                        & 0.6195 & 0.9476 & 0.2527 & 0.4586 & 1.9405 & 0.2721 \\
    \rowcolor{black!5} VBD     & 0.2276 & 0.4711 & 0.0819 & 0.2041 & 1.1579 & 0.2975 \\
    SMART                      & 0.1419 & 0.3976 & 0.0125 & 0.0617 & 0.8021 & 0.3510 \\
    \rowcolor{black!5} CATK    & 0.1426 & 0.4029 & 0.0095 & 0.0501 & 0.7988 & 0.3645 \\
    TrajTok                    & 0.1407 & 0.4173 & 0.0099 & 0.0524 & 0.8266 & 0.3652 \\
    \bottomrule
  \end{tabular}}
\end{table}

\subsection{Analysis on Replan Rate}

\begin{figure}[t]
  \centering
  \vspace{-10pt}
  \includegraphics[width=\linewidth]{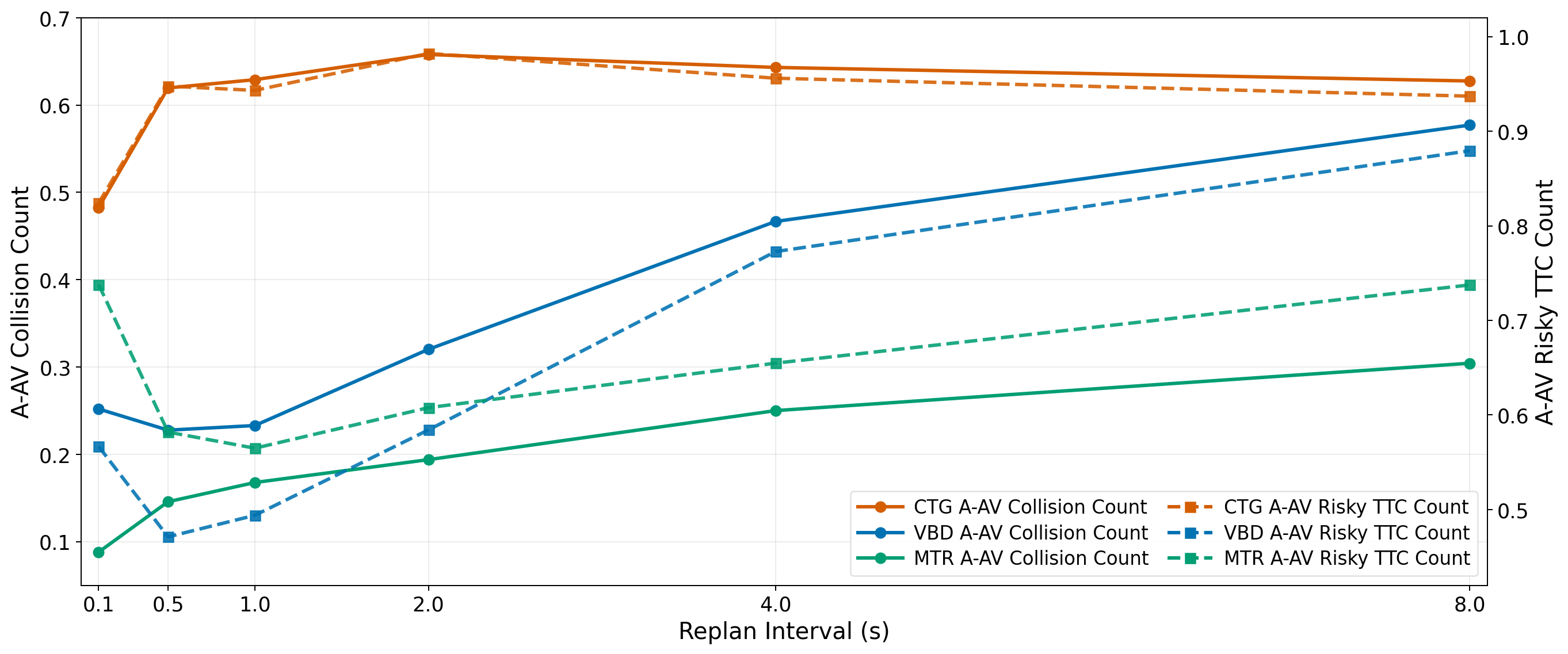}
  \caption{\textbf{Effect of replanning interval.} We ablate the replanning interval in ReactSim-Bench and report A-AV collision count and risky TTC count. More frequent replanning generally improves reactive capability by letting the simulator observe the realized AV behavior more often, but the effect remains model-dependent.}
  \label{fig:replan}
\end{figure}

Figure~\ref{fig:replan} ablates the replanning interval in ReactSim-Bench. As mentioned in Section~\ref{sec:3.1}, a model can predict multiple steps at once and then execute them alternately with the AV policy. The results show that higher replanning rates (above 1~Hz) generally bring better reactive capability. This is because the simulator does not know the future AV behavior in ReactSim-Bench, and more frequent replanning allows it to observe the current scene state, especially the realized AV behavior, more often and respond more effectively. However, frequent replanning can also introduce larger accumulated errors, so smaller replanning intervals are not always better for every model. Therefore, the replanning strategy can strongly affect reactive performance.

This differs from realism benchmarks such as WOSAC. In prior studies~\cite{lin2025revisit}, smaller replanning rates on WOSAC (below 0.5~Hz) have better realism. This is because WOSAC lets the model control all vehicles, including the AV, and does not consider interactions with vehicles outside model control. Therefore, accumulated error becomes the main negative effect of high-frequency replanning. To maintain the autoregressive requirement of simulation, WOSAC restricts the replanning rate to be greater than 1~Hz. \textbf{ReactSim-Bench does not need this type of restriction. Its demand for high-frequency replanning is natural and closer to practical simulation.}



\section{Conclusion}

In this work, we introduce ReactSim-Bench for evaluating the reactive capability of behavior world model simulation in autonomous driving. We decouple AV control from surrounding-agent simulation and collect deviated AV behaviors as benchmark inputs, allowing the simulator to be tested under situations where the AV does not simply follow the recorded log. We evaluate multiple state-of-the-art behavior world models under this protocol and provide a foundation and insights for developing future reactive behavior world models.

{
\small
\bibliographystyle{unsrtnat}
\bibliography{main}
}

\newpage
\appendix
\section{Case Study}
\label{app:case_study}

\begin{figure}[t]
  \centering
  \includegraphics[width=0.8\linewidth]{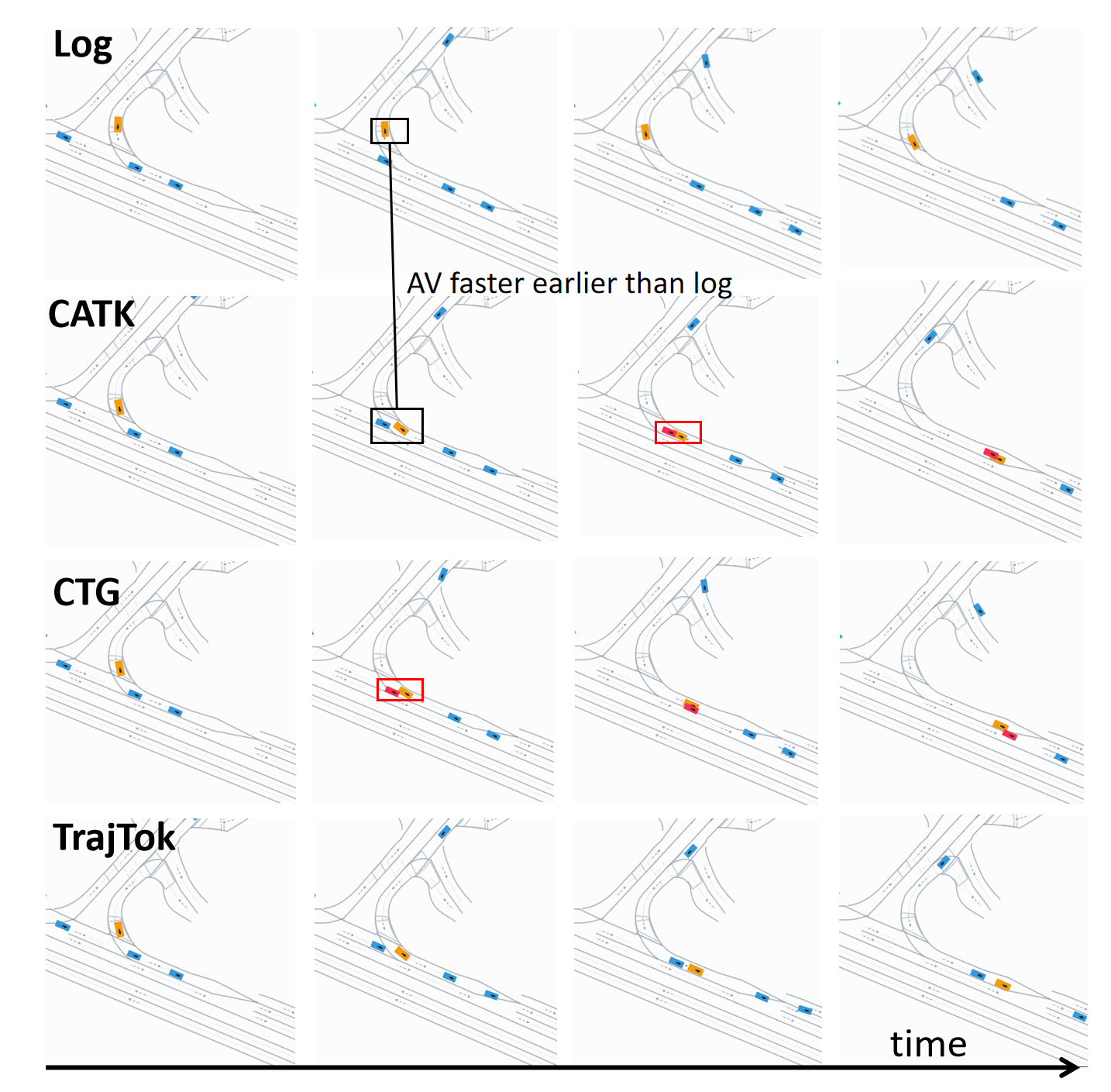}
  \caption{\textbf{Qualitative case study.} In a ramp-merging scenario, the deviated AV enters the main road earlier than in the log. CATK slows down but still collides, CTG remains close to log replay and collides, while TrajTok decelerates in time and avoids the collision.}
  \label{fig:vis}
\end{figure}

Figure~\ref{fig:vis} shows a merging scenario where the AV on the ramp starts earlier than in the log and merges onto the main road, requiring the rear vehicle to yield. CATK produces a deceleration response, but the deceleration is insufficient. CTG largely ignores the change in AV behavior, keeping the rear vehicle close to the logged trajectory and directly colliding with the AV. In contrast, TrajTok decelerates in time and avoids the potential collision. The two failure cases suggest that when the AV behavior differs from the log, simulators may still tend to make surrounding agents follow the recorded trajectories.

\begin{figure}[t]
  \centering
  \includegraphics[width=0.85\linewidth]{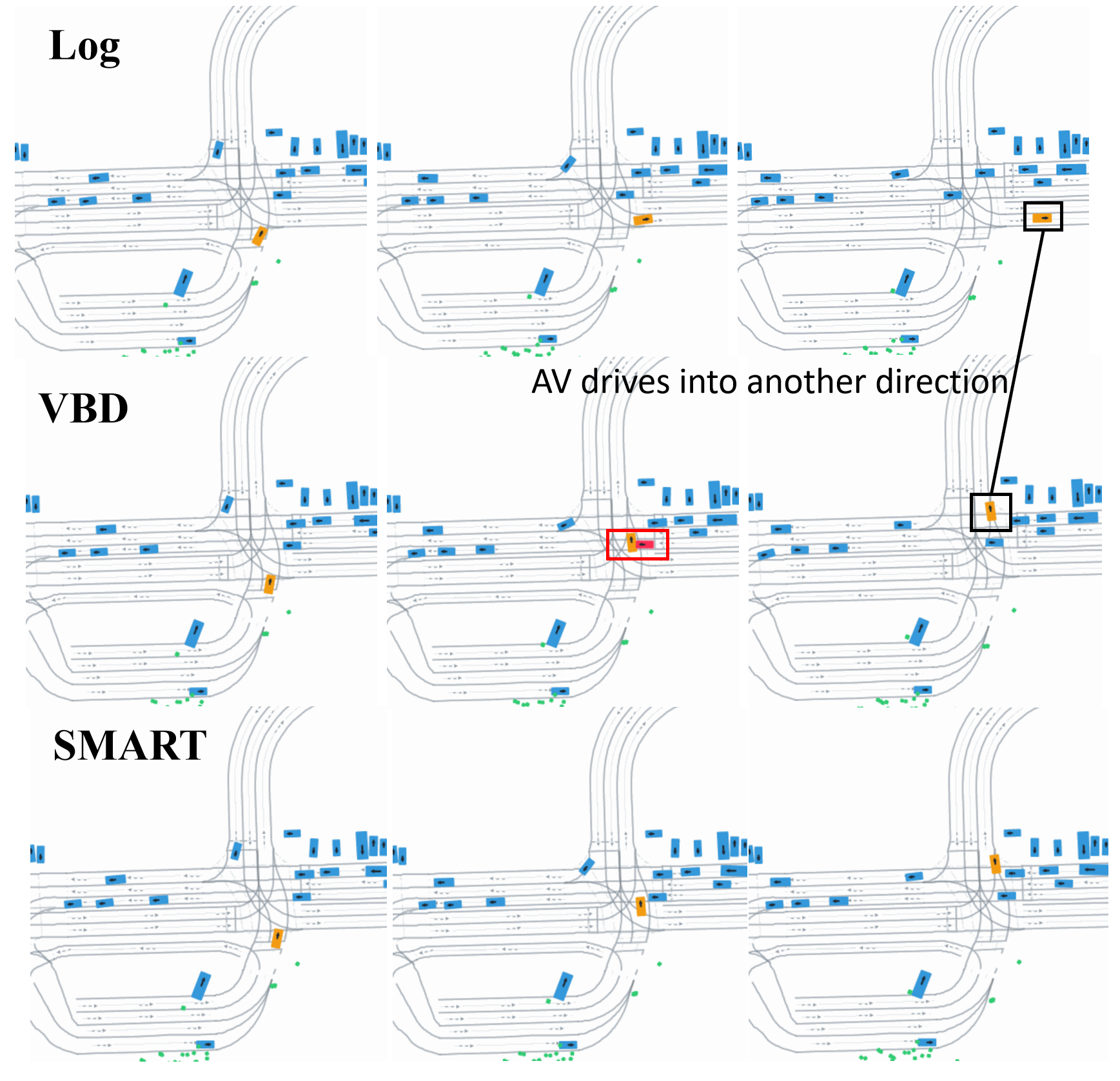}
  \caption{\textbf{Directional-deviation case study.} The collected AV input drives toward a direction different from the log. VBD keeps the interacting vehicle close to the logged behavior and collides with the AV, while SMART produces a safer response.}
  \label{fig:vis_direction}
\end{figure}

Figure~\ref{fig:vis_direction} shows another case with a directional deviation. In the logged behavior, the AV follows the original traffic flow, while the collected AV input drives toward a different direction and creates a new interaction with nearby vehicles. VBD does not sufficiently adapt the surrounding-agent behavior to the deviated AV motion. The interacting vehicle remains close to its logged trajectory and collides with the AV, as highlighted in the red box. SMART produces a more compatible response and avoids the collision. 

\section{Limitations}

We mainly focus on evaluating the reactive capability of behavior world model simulation. In addition to reactivity and realism, some methods also target other objectives, such as controllability and diversity. A systematic benchmark for these objectives is outside the scope of this paper, but may serve as a direction for future research.

\section{Implementation Details}

\subsection{Baseline Implementation on nuPlan}

\textbf{Scenario Data.} All models are trained on the same unified per-scenario nuPlan data cache. Our current cache contains 642,639 training scenarios and 2,636 validation scenarios. Each scenario is represented by 91 frames sampled at 10~Hz, with the current frame fixed at index 10, yielding 11 history frames and 80 future frames. We keep the ego vehicle plus up to 63 nearby agents in each scene, using the same current-time index and the same train/validation split for all methods. Agent state tensors contain position, heading, velocity, and object size, and are padded to the same history and future lengths before model-specific preprocessing.

We first map raw nuPlan tracked-object types into a canonical agent taxonomy. Specifically, \texttt{VEHICLE} and \texttt{EGO} are mapped to the vehicle class, \texttt{PEDESTRIAN} to the pedestrian class, and \texttt{BICYCLE} to the cyclist class, while \texttt{TRAFFIC\_CONE}, \texttt{BARRIER}, \texttt{CZONE\_SIGN}, and \texttt{GENERIC\_OBJECT} are merged into an auxiliary ``other'' category in the cache. For training the world models, we use the three shared dynamic classes vehicle / pedestrian / cyclist, and ego is always preserved as slot 0. The remaining slots are filled by selecting the nearest valid agents around ego at the current frame, up to the shared limit of 64 total agents. This gives all methods the same scene-centric agent budget and the same temporal alignment before their own feature builders are applied.

The unified cache is built from original nuPlan map API queries around the scenario region. During preprocessing, we collect lane, lane connector, crosswalk, and stop-line map objects, and derive road-edge geometry from lane and lane-connector boundaries. In the common representation used across methods, map elements are organized into four semantic categories: lane, road edge, stop line, and crosswalk. Each element is represented as a sampled polyline together with a validity mask. Lane connectors are merged into the lane class in the shared vector representation, but connector-specific identity is retained in the cache when downstream modules need exact traffic-light alignment. Traffic lights are stored through stop-point coordinates, future state sequences, and lane-polyline indices, with raw nuPlan states normalized into a canonical state space before model-specific remapping. This common cache is then adapted into each model's native input format, so that differences across methods mainly come from the model architecture rather than scene parsing or map extraction.

\textbf{Model Settings.} All models are trained on 8 NVIDIA H200 GPUs. We preserve the original optimizer family and the original model architecture of each baseline whenever possible, and only replace the dataset interface and nuPlan alignment layer required to consume the unified data cache. The model-specific settings are summarized in Table~\ref{tab:appendix_model_settings}.

\begin{table}[h]
    \centering
    \caption{Model-specific training and map settings under the unified nuPlan data cache.}
    \label{tab:appendix_model_settings}
    \scriptsize
    \setlength{\tabcolsep}{5pt}
    \resizebox{0.99\linewidth}{!}{
    \begin{tabular}{p{1.45cm}p{3.3cm}p{3.2cm}p{1.15cm}p{1.65cm}p{1.2cm}}
    \toprule
    Model & Architecture & Map setting & Batch Size & Training & LR \\
    \midrule
    VBD & 6-layer encoder; 50 diffusion steps; 2-step actions & 300~m crop; 256 polylines; 30 pts / polyline & 32 scenarios & 16 epochs & $2\times10^{-4}$ \\
    MTR & 6-layer encoder; 6-layer decoder; 6 modes; hidden 256 / 512 & 120~m radius; 768 source polylines; 20 pts / polyline & 8 scenarios & 12 epochs & $10^{-4}$ \\
    CTG & ResNet-18 BC policy & $224\times224$ raster; 0.5~m / px; $112\times112$~m & 1,600 agents & 1M iters & $10^{-3}$ \\
    SMART & hidden dim 128; 6 agent layers & 300~m crop; tokenized vector map & 10 scenarios & 32 epochs & $5\times10^{-4}$ \\
    TrajTok & hidden dim 128; 6 agent layers; new traj. vocab; LS + SAS & 300~m crop; tokenized vector map & 10 scenarios & 32 epochs & $5\times10^{-4}$ \\
    CLSFT & SMART init.; closed-loop SFT & 300~m crop; tokenized vector map & 10 scenarios & fine-tuning 32 epochs on SMART & $5\times10^{-5}$ \\
    \bottomrule
    \end{tabular}}
\end{table}

\subsection{Metric Computation Details}
\label{app:metric_details}

This section provides the implementation details of the seven metrics used in ReactSim-Bench. Let $\mathcal{V}$ denote the set of evaluated non-ego vehicles, $\mathcal{A}$ denote the set of evaluated non-ego objects, and $T$ denote the number of future simulation steps. All scenarios are sampled at $\Delta t=0.1$~s. For metrics reported as rates, we first compute a scenario-level rate and then average it over all valid scenarios. For metrics reported as counts, the reported value is the mean scenario-level count. Collision-related metrics use oriented bounding boxes in bird's-eye view. A pair of boxes is treated as colliding when their rounded-box signed distance is below $0$.

\paragraph{Agent-AV collision count.}
This metric counts at-fault collision events between evaluated non-ego vehicles and the AV. For each scenario, we first identify new collision events between a vehicle $i\in\mathcal{V}$ and the AV, where a repeated overlap between the same pair is counted only at its first collision step. A new collision event is counted only if the non-ego vehicle is judged to be at fault. Specifically, the event is ignored if the non-ego vehicle is stopped or if the AV is behind the non-ego vehicle. It is counted as at-fault if the AV is stopped, if the front bumper segment of the non-ego vehicle intersects the AV box, or if the non-ego vehicle is already off-road at the collision step. The scene-level metric is
\begin{equation}
\mathrm{AAVColl}
= \sum_{i\in\mathcal{V}} \mathds{1}\{\exists t:\mathrm{NewColl}_{i,\mathrm{AV}}(t)\land \mathrm{AtFault}_{i,\mathrm{AV}}(t)\}.
\end{equation}
In the implementation, the stopped-speed threshold is $0.05$~m/s, and the rear-collision test uses a $150^\circ$ angular threshold.

\paragraph{Agent-AV risky TTC count.}
This metric counts how many evaluated non-ego vehicles produce a risky time-to-collision (TTC) with the AV. For each vehicle, the AV is considered a possible front object only when it lies ahead of the vehicle, has sufficient lateral overlap, and has a heading difference within the front-object filtering threshold. The relative closing speed is computed as $v_i-v_{\mathrm{AV}}$. If the pair is valid and the closing speed is positive, we compute
\begin{equation}
\mathrm{TTC}_{i,\mathrm{AV}}(t)
= \min\left(\frac{d^{\mathrm{long}}_{i,\mathrm{AV}}(t)}{v_i(t)-v_{\mathrm{AV}}(t)}, 5.0\right),
\end{equation}
where $d^{\mathrm{long}}_{i,\mathrm{AV}}$ is the longitudinal gap from vehicle $i$ to the AV after accounting for vehicle extents. If the boxes already overlap, the TTC is set to $0$. The scene-level metric is
\begin{equation}
\mathrm{AAVTTC}
= \sum_{i\in\mathcal{V}}\mathds{1}\{\min_t \mathrm{TTC}_{i,\mathrm{AV}}(t)<0.5~\mathrm{s}\}.
\end{equation}

\paragraph{Agent-agent collision rate.}
This metric measures how often evaluated non-ego objects collide with other objects. At each future step, we compute the minimum signed distance between each evaluated object $i\in\mathcal{A}$ and all other valid objects. Self-pairs and invalid pairs are masked out. The implementation also excludes pedestrian--pedestrian, pedestrian--cyclist, and cyclist--pedestrian pairs from this metric. Let $C_i(t)$ indicate whether object $i$ overlaps with any remaining valid object at step $t$. The scene-level metric is
\begin{equation}
\mathrm{AACollRate}
= \frac{1}{|\Omega_{\mathrm{coll}}|}
\sum_{(i,t)\in\Omega_{\mathrm{coll}}} C_i(t),
\end{equation}
where $\Omega_{\mathrm{coll}}$ is the set of valid evaluated object-step pairs.

\paragraph{Off-road rate.}
This metric measures whether non-ego vehicles leave the drivable area. For each vehicle and future step, we query the nuPlan map and mark the step as off-road if the vehicle center point $(x_i(t),y_i(t))$ is not inside the \texttt{DRIVABLE\_AREA} layer. Vehicles whose center point is already off-road at the current frame are excluded from the denominator. The scene-level metric is
\begin{equation}
\mathrm{OffroadRate}
= \frac{1}{|\Omega_{\mathrm{off}}|}
\sum_{(i,t)\in\Omega_{\mathrm{off}}}
\mathds{1}\{(x_i(t),y_i(t))\notin \mathrm{DRIVABLE\_AREA}\}.
\end{equation}
This implementation checks the vehicle center point rather than all corners of the bounding box.

\paragraph{Driving-direction violation rate.}
This metric is an agent-level rate over non-ego vehicles. For each vehicle, we first project valid poses onto the corresponding nuPlan route baseline and compute progress along the lane centerline. A raw wrong-way violation at time $t$ requires the vehicle to move faster than $1.0$~m/s, have a heading error larger than $75^\circ$ relative to the lane direction, and accumulate more than $2.0$~m of negative progress over a $1.0$~s window. A raw violation becomes a final violation only if it persists for at least $0.5$~s. The scene-level metric is
\begin{equation}
\mathrm{DirRate}
= \frac{1}{|\mathcal{V}_{\mathrm{dir}}|}
\sum_{i\in\mathcal{V}_{\mathrm{dir}}}
\mathds{1}\{\exists t:\mathrm{PersistWrongWay}_i(t)\},
\end{equation}
where $\mathcal{V}_{\mathrm{dir}}$ contains vehicles with at least one valid candidate step.

\paragraph{Acceleration infeasibility rate.}
Following the Waymax kinematic infeasibility metric~\cite{gulino2023waymax}, our two kinematic metrics estimate acceleration and steering curvature from consecutive vehicle states and use the same bounds, i.e., acceleration magnitude below $6.0$~m/s$^2$ and steering-curvature magnitude below $0.3$~m$^{-1}$. We report the two violations separately as step-level rates rather than as a single kinematic-infeasibility flag.
This metric measures the fraction of valid non-ego vehicle transitions that violate the acceleration bound. Velocities are taken from the trajectory state when available; otherwise they are computed from position differences. We concatenate the last history frame with the future rollout so that the first transition from history to prediction is also checked. For vehicle $i$ and transition $t$,
\begin{equation}
a_i(t)=\frac{\|v_i(t+1)\|_2-\|v_i(t)\|_2}{\Delta t}.
\end{equation}
The violation indicator is $\mathds{1}\{|a_i(t)|>6.0+10^{-3}\}$, and the scene-level metric is the mean of this indicator over valid non-ego vehicle transitions.

\paragraph{Steering-curvature infeasibility rate.}
This metric measures the fraction of valid non-ego vehicle transitions that violate the steering-curvature bound. Let $\Delta\psi_i(t)$ be the wrapped heading change and let
\begin{equation}
d_i(t)=\|v_i(t)\|_2\Delta t+\frac{1}{2}a_i(t)\Delta t^2
\end{equation}
be the approximate traveled distance over the transition. The steering curvature is computed as
\begin{equation}
\kappa_i(t)=\frac{\Delta\psi_i(t)}{d_i(t)}.
\end{equation}
When the current or next speed is below $0.6$~m/s, the curvature is set to $0$ to avoid unstable low-speed estimates. The violation indicator is $\mathds{1}\{|\kappa_i(t)|>0.3+10^{-3}\}$, and the scene-level metric is the mean of this indicator over valid non-ego vehicle transitions.

\subsection{Details about data collection}

We use Diffusion Planner as the AV planner to collect data that differ from the log. We use the official open-source model weights to perform inference on nuPlan scenarios. Following the original paper, we remove navigation information and increase the temperature to improve its multimodal capability. Diffusion Planner generates only one trajectory per inference. For each scenario, we independently repeat inference 16 times and obtain diverse trajectories through stochastic sampling in diffusion.

We build an interactive data verification tool that visualizes the scenario and candidate AV behaviors and displays information such as speed and TTC, making it easier for annotators to select candidates and verify whether the filtered AV behaviors satisfy our requirements.



\end{document}